# A Lightweight Stemmer for Gujarati

Juhi Ameta, Nisheeth Joshi, Iti Mathur

Department of Computer Science, Apaji Institute, Banasthali University, Rajasthan, India,

*juhi_214@yahoo.co.in, nisheeth.joshi@rediffmail.com, mathur_iti@rediffmail.com*

**ABSTRACT**

*Gujarati is a resource poor language with almost no language processing tools being available. In this paper we have shown an implementation of a rule based stemmer of Gujarati. We have shown the creation of rules for stemming and the richness in morphology that Gujarati possesses. We have also evaluated our results by verifying it with a human expert.*

## 1. INTRODUCTION

Stemming is a process in which suffixes are removed from its root word or stem. This relates morphologically variant words to its common root. For example: authored, authorized, authorship are morphological variants of the word 'author'. The process of grouping morphemically different words to a common root is termed as conflation. It is an important tool which support in the development of various natural language processing (NLP) applications. To name a few, it can be used in information extraction, search engines, machine translation evaluation, spell checking etc.

Stemmers are language specific tools. The design of a stemming algorithm requires a significant level of linguistic expertise. There has been a lot of significant work in the development and evaluation of stemmers for European languages, but very less or no significant work has been done on Indian front. For a language like Gujarati, the problem engraved even further as we have very less language resource and tools for the development of the same. With the advent of online information being available in Gujarati, a need was felt to develop tools for Gujarati language processing. This paper is part of an ongoing research where we're developing an information retrieval system for Gujarati.

In this paper we describe the development process of lightweight Gujarati stemmer which strips the suffixes based on the longest match. For Indian languages, this approach was first suggested by Ramanathan and Rao [1] who performed the same activity onto Hindi. This approach is very robust and domain independent. The rest of the paper is as follows: In section 2 we review the work done in development of stemmer in popular European and Indian Languages. In section 3; we provide a brief description of Guajarati language. In section 4, we explain the implementation of our proposed stemming algorithm. In section 5, we evaluate the performance of our stemmer. Finally, in section 6, we conclude our work with some suggestions for further enhancements.

## 2. RELATED WORK

Very first stemming algorithm for English was given by Lovins [2] in 1969. This paper started the debate of using stemming algorithms in NLP applications. During early seventies two more stemming algorithms were proposed by Hafer and Wsiss [3] and Paice [4]. In 1980, Martin Porter proposed a suffix stripping algorithm [5]. In the years to come, this algorithm became the standard stemming algorithm. This algorithm was first proposed for English and later on was extended to other European languages .Today mostly all the NLP systems for European languages, which require a stemmer use porter's algorithm. Another approach to stemming was proposed by Frakes and Baeza-Yates [6] who proposed the use of term indexes and its root word in a table lookup.

Since the dawn of the new century, as the processing and storage capabilities improved the stemmers saw a paradigm shift from a rule based approach to statistical/machine learning approaches. Some of the most prominent work in this area have been done by Goldsmith [7][8] who proposed an algorithm to model morphological variants of European languages using an unsupervised approach. Crutz [9] proposed an unsupervised segmentation approach using maximum posteriori. Sover and Bent [10] proposed a Bayesian model for stemming of English and French languages. Freitag [11] proposed an algorithm for clustering of words using co-occurrence information.

In an Indian context, the first efforts to implement a stemmer was done by Larsky et al [12] who used a rule based approach and 27 rules to implement a stemmer for Hindi. Ramanathan and Rao [1] used the same approach, but used some more rules for stemming. Dasgupta and ng [13] proposed an unsupervised morphological stemmer for

Bengali. Majumdar et al. [14] proposed a cluster based approach based on string distance measures which required no linguistic knowledge. Pandey and Siddiqui [15] proposed an unsupervised approach to stemming for Hindi, which was mostly based on the work of Goldsmith [7].

### 3. GUJARATI LANGUAGE

Gujarati is a highly inflectional language with a relatively free word order. It has three genders (masculine, neuter and feminine). Nouns, verbs, adjectives and adverbs can have multiple inflectional forms. Verbs can have gender, number, person, tense, aspect and modality based infected forms. Nouns can have nominative, objective and locative cases. Moreover, unlike Hindi and most of the Indian languages, Gujarati can have multiple postpositions added with the nouns or verbs. For Examples: ગુજરાતમાં, ગુજરાતનું, ગુજરાતનો, ગુજરાતમાંથી, ગુજરાતની etc can be added with ગુજરાત, creating different variants of the word. In the following subsection, a brief explanation is provided on the usage of open classes (noun, verb, adjective, adverbs) in Gujarati language.

#### 3.1 Nouns:

There are three genders in Gujarati as in Sanskrit, namely – masculine, feminine and neuter. Numbers as in other languages are singular and plural. A noun can be represented as noun stem+gender marker+number marker. If traditional grammar is followed then for some words, the word is stemmed wrongly as in પાણી(paan +i) whereas it should be treated as a complete stem પાણી. Thus, Gujarati has many nouns which end in –o and –i which don't fall under the category of masculine or feminine respectively. Table 1 provides a brief description of nouns.

| Gender | Singular | Plural |
|---|---|---|
| Feminine | છોકરી (chhokar+i) (girl) | છોકરીઓ (chhokar+i+o) (girls) |
| Masculine | છોકરો (chhokar+o) (boy) | છોકરાઓ (chhokar+aa+o) (boys) |
| Neuter | છોકરું (chhokar+uN) (child) | છોકરાંઓ (chhokar+aaN+o) (children) |

Table 1: Nouns in Gujarati

#### 3.2. Verbs

Verbs may be inflected or non-inflected. The non-inflected forms have the form - verb stem + infinitive. For example, કાપવું (કાપ+વું) (kaap+vuN) (to cut). Verbs may be simple as in હસવું (has+vuN) or derived as in હસાવવું (has+aavvuN). Inflected verbs have the form verb stem + inflectional material. Some of the inflections possible for the verb, like રડવું (rad+vuN) (to cry), are described in tables 2, 3, 4 and 5.

| Person | Singular | Plural |
|---|---|---|
| First Person | રડું (rad+uN) | રડીએ (rad+ie) |
| Second Person | રડે (rad+e) | રડો (rad+o) |
| Third Person | રડે (rad+e) | રડે (rad+e) |

Table 2: Verbs with Present Tense

| Person | Singular | Plural |
|---|---|---|
| First Person | રડીશ (rad+ish) | રડીશું (rad+ishuN) |
| Second Person | રડશે (rad+she) | રડશો (rad+sho) |
| Third Person | રડશે (rad+she) | રડશે (rad+she) |

Table 3: Verbs with Future Tense

| Gender | Singular | Plural |
|---|---|---|
| Masculine | રડતો (rad+to) | રડતા (rad+taa) |
| Feminine | રડતી (rad+ti) | રડતી (rad+ti) |
| Neuter | રડતું (rad+tuN) | રડતાં (rad+taaN) |

Table 4: Verbs with Past Progressive

| Gender | Singular | Plural |
|---|---|---|
| Masculine | રડ્યો (rad+yo) | રડ્યા (rad+yaa) |
| Feminine | રડી (rad+i) | રડી (rad+i) |
| Neuter | રડ્યું (rad+yuN) | રડ્યાં (rad+yaaN) |

Table 5: Verbs with Past Perfect Tense

#### 3.3 Adjectives

Adjectives may either be variable or invariable. The variable adjectives change with person and gender. Consider the example સારું (saar+o) meaning good:

| Gender | Singular | Plural |
|---|---|---|
| Masculine | સારો (saar+o) | સારા (saar+a) |
| Feminine | સારી (saar+i) | સારી (saar+i) |
| Neuter | સારું (saar+uN) | સારાં (saar+aaN) |

Table 6: Various forms of word સારું

On the other hand the invariable adjectives never change with a change in person or gender.

For example, સુંદરછોકરો – sundar chhokar+o (a beautiful boy)

સુંદરછોકરી – sundar chhokar+i (a beautiful girl)

સુંદરછોકરું – sundar chhokar + uN (a beautiful child)

સુંદરછોકરીઓ – sundar chhokar+io (beautiful girls)

### 3.4 Adverbs

Adverbs may be variable or invariable just like adjectives, depending upon whether they change the form with the noun with which the verb agrees. Table 7 categorizes some of the adverbs.

| Category | Example of adverb |
|---|---|
| Time | આજે (aaje) (today) |
| Place | અહીં (ahiN) (here) |
| Manner | અચાનક (achaanak) (suddenly) |
| Order | પહેલાં (pahelaaN) (before) |
| Quantity | ઘણું (ghanuN) (much) |
| Doubt | ક્યારેક (kyaarek) (occasionally) |
| Frequency | રોજ (roj) (everyday) |
| Negative | ના (naa) (no) |
| Connecting | પરિણામે (parinaame) (consequently) |

Table 7: Some Categories of Adverbs

### 4. PROPOSED SYSTEM

In order to create a stemmer for Gujarati we created a suffix list. A suggestive list is shown in Figure 1. In our algorithm we removed longest possible suffix from the list. In all we used 167 suffixes for extraction of root words.

| | | | | |
|---|---|---|---|---|
| ાઓમાનું | ાઓનાં | ીમાંથી | ીઓને | ીમાંથી |
| ાચેલો | માંની | ાઓ | ાય | ીશ |
| સ્વી | િક | થી | વી | તું |
| ના | ો | ી | ે | ા |

Figure 1: Suggestive Suffix List

We do not claim that this is an exhaustive list. Some more suffixes could be added, but in our experiments we found that on adding more suffixes our stemmer started to over-stem the words, causing much greater overstemming errors. So, we did not add any more suffix in our approach. These suffixes were ordered according to the length, from longest to the shortest sequence. When an input word is given it is stripped based on the suffix list. The larger ones are removed first and then if required shorter ones are removed and so on. In some of the cases we could not find the right suffixes. For example: word સેવાનો should have been reduced to સેવા but we actually got સે. Table 8 Shows some of the stems generated by our algorithm.

| Word | Stem | Suffix |
|---|---|---|
| શહેરી | શહેર | ી |
| વિસ્તારોમાં | વિસ્તાર | ોમાં |
| ભાજપનો | ભાજપ | નો |
| સફાયો | સફ | ાયો |
| દેશને | દેશ | ને |
| બુટાસિંહને | બુટાસિંહ | ને |
| અદાલતને | અદાલત | ને |
| અસીલોએ | અસીલ | ોએ |
| વકીલોની | વકીલ | ોની |

Table 8: Result of the proposed stemming algorithm

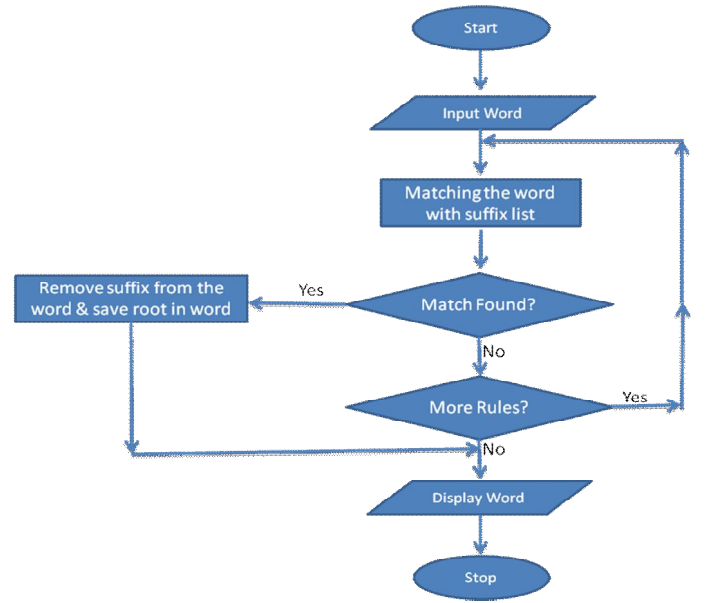

Figure 2: Lightweight stemmer for Gujarati

Our approach successfully captures most of the morphological variations of a word with a good accuracy. The result of the evaluation are shown in the section next section. Figure 2 shows the working of our stemming algorithm.

## 5. EVALUATION AND RESULTS

To evaluate our system, we registered the no. of times the system committed under stemming errors (suffixes which were supposed to be removed but were not removed) and over stemming errors (suffixes which were not supposed to be removed but are removed). In order to perform this analysis we created a test corpus of 3000 words. Table 9 shows the summary of the test data.

| Test Data Features | Total Count |
|---|---|
| Total Words | 3000 |
| Unique Words | 389 |
| Stem Groups with more than one words | 218 |
| Stem Groups with only one words | 171 |
| Min Length | 2 |
| Max Length | 17 |

Table 9: Summary of Test Data

In this 3000 words test data, we had 389 words which were unique i.e. we had 389 stems in the entire test corpus. Among these 389 words 218 stems had more than one morphological variants present in the corpus and 171 stems had only one morphological variant.

We executed our algorithm on this test data and stored the results in the format shown in Table 8. A human was asked to manually check if the roots and stems generated are correct or not. It was found that, out of the 3000 words 255 times the system could not provide the correct results. This gave us the accuracy of 91.5%. Out of the 255 errors, 189 were over-stemming errors and 36 were under-stemming errors. Thus, 86% of the total errors were due to over-stemming and 14% of the total errors were due to under-stemming.

## 6. CONCLUSION AND FUTURE WORK

We have shown the design and implementation of rule based stemmer for Gujarati. The stemmer is able to capture most of the morphological variants. We tested our systems for the verification of our claim and have achieved 91.5% accuracy in the same.

As an extension to this work, we would like to perform a more rigorous error analysis so that a detailed error analysis report can be provided. Moreover our approach is more prone to over-stemming errors, in future we would also would like to reduce these errors, thus improving the accuracy ever further.